**Application of 2D Homography for High Resolution Traffic Data Collection using CCTV Cameras**

**Linlin Zhang**
PhD Student
Department of Civil and Environmental Engineering
University of Missouri-Columbia, Columbia, MO, USA, 65201
Email: lz5f2@mail.missouri.edu

**Xiang Yu, P.E.**
PhD Student
Department of Civil and Environmental Engineering
University of Missouri-Columbia, Columbia, MO, USA, 65201
Email: xytm4@mail.missouri.edu

**Abdulateef Daud**
Traffic EIT/Coordinator
HDR
401 South 18th Street, Suite 300, St. Louis, MO, USA, 63103
Email: aadcvg@umsystem.edu

**Abdul Rashid Mussah**
PhD Student
Department of Civil and Environmental Engineering
University of Missouri-Columbia, Columbia, MO, USA, 65201
Email: akm2fx@mail.missouri.edu

**Yaw, Adu-Gyamfi**
Associate Professor
Department of Civil and Environmental Engineering
University of Missouri-Columbia, Columbia, MO, USA, 65201
Email: adugyamfiy@missouri.edu

Word Count: 6948 words + 7 tables (250 words per table) = 8699 words

**ABSTRACT**

Traffic cameras remain the primary source data for surveillance activities such as congestion and incident monitoring. To date, State agencies continue to rely on manual effort to extract data from networked cameras due to limitations of the current automatic vision systems including requirements for complex camera calibration and inability to generate high resolution data. This study implements a three-stage video analytics framework for extracting high-resolution traffic data such vehicle counts, speed, and acceleration from infrastructure-mounted CCTV cameras. The key components of the framework include object recognition, perspective transformation, and vehicle trajectory reconstruction for traffic data collection. First, a state-of-the-art vehicle recognition model is implemented to detect and classify vehicles. Next, to correct for camera distortion and reduce partial occlusion, an algorithm inspired by two-point linear perspective is utilized to extracts the region of interest (ROI) automatically, while a 2D homography technique transforms the CCTV view to bird's-eye view (BEV). Cameras are calibrated with a two-layer matrix system to enable the extraction of speed and acceleration by converting image coordinates to real-world measurements. Individual vehicle trajectories are constructed and compared in BEV using two time-space-feature-based object trackers, namely Motpy and BYTETrack. The results of the current study showed about +/- 4.5% error rate for directional traffic counts, less than 10% MSE for speed bias between camera estimates in comparison to estimates from probe data sources. Extracting high-resolution data from traffic cameras has several implications, ranging from improvements in traffic management and identify dangerous driving behavior, high-risk areas for accidents, and other safety concerns, enabling proactive measures to reduce accidents and fatalities.

**INTRODUCTION**

Digital Twins - a bidirectional flow of intelligence between the virtual and the physical world - is one of the most promising technologies that will benefit the ITS industry tremendously. When provided with accurate, continuous, high-resolution traffic data in the real world, the promise of DT is to give deeper insights into the intricacies of traffic flow, the interaction between the different components of the system, and look further into the future behavior of transportation system under various conditions. Therefore, one of the challenges posed by transportation agencies is the acquisition of spatial-temporally continuous high-resolution data that will be able to power this DT technology. Early researchers have used infrastructure-mounted sensors such as loop detectors, radar-based sensors, and GPS-equipped vehicles for traffic data collection. However, these conventional methods are gradually being replaced by more efficient and advanced emerging technologies such as CCTV cameras, Unmanned Aerial Vehicles (UAVs), etc. for numerous reasons, including high installation and maintenance costs, disruptions to traffic flow during installations, low penetration, and privacy issues in GPS-based data collection. Ultimately, the goal of the current research effort is to leverage the proliferation of these advanced, cost-effective technologies and the recent advancement in artificial intelligence to develop a robust, end-to-end methodological framework that will be able to extract high-resolution traffic data that will drive intelligent transportation system into a safe and sustainable future.

Numerous techniques for collecting high-resolution traffic data have been discussed in the literature. Early researchers (Bennett et al., 2006; Minge, 2010; Sheik Mohammed Ali et al., 2012) explored the use of both intrusive and non-intrusive sensors such as inductive-loop detectors and radar-based devices for the collection of granular vehicular data such as vehicle speed, classification, headway, vehicle counts (Minge, 2010). These sensors remain the most popular technique for traffic data collection due to their ability to collect data in all types of weather conditions and at night. They are however not suitable for trajectory construction due to spatial limitations and therefore not suitable for corridor level analysis. A more promising alternative is connected car data which provides rich vehicle trajectory information streaming directly from vehicles on various highways. (Altintasi et al., 2017; Llorca et al., 2010; Treiber and Kesting, 2013)have also explored the advantages of this dataset including high fidelity individual speeds, and acceleration, braking events, etc. Unfortunately, due to low penetration rates, traffic flow variables such as flow rates, density, gaps cannot be estimated from this technology currently.

In the pursuit of more comprehensive traffic data, advancements in video surveillance technology have offered another avenue for gathering high-resolution metrics. Video cameras, often augmented by computer vision algorithms, can capture not just speed, classification, and counts, but also more complex behaviors like lane-changing, overtaking, and even driver compliance with traffic signals. However, video data collection is not without its challenges. The technology is sensitive to lighting conditions, requiring good illumination for optimal performance. Furthermore, computational requirements for real-time analysis can be intensive, and there are also concerns about privacy and data protection. In addition to hardware-based solutions, advances in machine learning and data analytics have ushered in a new era of traffic data collection and analysis. Machine learning algorithms can be trained to analyze data from multiple sources, including sensors, cameras, and connected cars, to create more accurate and holistic traffic models. These can predict traffic flow, identify congestion patterns, and even suggest mitigative measures. However, these techniques require large and diverse datasets for training, and their effectiveness is constrained by the quality and granularity of the data they are trained on.

While traditional intrusive and non-intrusive sensors like inductive-loop detectors and radar-based devices have their limitations, they are still widely used for their robustness in varying conditions. Newer technologies like video surveillance, LiDAR, and machine learning offer increased granularity and capabilities but come with their own sets of challenges (Bourja et al., 2021; Gargoum and El Basyouny, 2019). Connected car data remains an exciting frontier but is currently limited by low penetration rates (Mussah and Adu-Gyamfi, 2022). The objective of this study thus looks to augment the current efforts of inductive-loop detectors and radar-based data collection with defined video analytics techniques. By fusing these diverse sets of data, we aim to overcome the individual limitations of each technology to create a more robust, comprehensive, and high-resolution traffic monitoring system. The integration of video analytics allows for the capture of complex traffic behaviors that are not easily obtained through inductive-loop detectors or radar alone, such as lane-changing dynamics, compliance with traffic laws, and even pedestrian interactions. This multi-modal approach seeks not only to improve the granularity of the data collected but also to provide a more holistic understanding of traffic flow and congestion patterns. Through this fusion of technologies, the study aims to enhance the quality of traffic management and planning, thereby contributing to safer and more efficient transportation networks.

**LITERATURE REVIEW**

Early work in traffic monitoring primarily relied on intrusive sensors like inductive-loop detectors, which have been extensively studied and deployed for their ability to gather basic metrics such as vehicle counts, speed, and headway. They have been lauded for their robustness, functioning reliably in varied weather conditions and during nighttime. Similarly, non-intrusive radar-based sensors have gained prominence for their ease of installation and maintenance, as discussed in multiple studies. However, literature has consistently pointed out that both these technologies are limited in their spatial resolution, making them less suitable for detailed corridor-level analysis or trajectory-based studies. In recent years, attention has shifted towards more advanced methods of data collection. Research on video analytics, often enhanced by computer vision and machine learning algorithms, has shown promising results in capturing complex vehicular behaviors, albeit with limitations related to lighting conditions and computational overhead.

Several studies (Giannakeris et al., 2018; Kovvali et al., 2007; Park et al., 2017; Zhang and Jin, 2019) have also focused on using closed-circuit television (CCTV) cameras for the extraction of high-resolution vehicle data. Most US highways are equipped with pole-mounted CCTV for traffic surveillance. These cameras have tilt, pan, zoom (TPZ) functionality that enable them to have more flexible and dynamic coverage of traffic scenes. (Park et al., 2017) leveraged this advancement in CCTV technology to propose a vision-based framework that can automatically extract traffic flow information from these videos. In the study, the camera's field of view was first calibrated to allow for real world measurements, followed by vehicle detection using background subtraction and Haar-Cascade. A combination of trackers was subsequently used to construct trajectories from detected vehicles for extraction of traffic flow variables. The authors reported about 7.5% error in traffic volume estimation while the accuracy of traffic speeds estimated were comparable to GPS based systems. The major drawback of this approach and many other CCTV-based vision systems (Javadi et al., 2019; Rodríguez-Rangel et al., 2022; Shirazi and Morris, 2016; Sochor et al., 2019) is the requirement to calibrate each camera for perspective distortion. Vehicle occlusion can also be a challenge that could affect the trajectory reconstruction process. The Next Generation Simulation Dataset (Minge, 2010) is arguably the most comprehensive study that leveraged CCTV camera data to collect high resolution traffic data for microsimulation. The study extracted individual vehicle trajectories from videos taken by multiple digital cameras installed on nearby building roofs covering multiple freeways of interest. Feature-based detection algorithm and zero-mean cross-correlation matching algorithm were used for vehicle detection and tracking, respectively. Although the NGSIM dataset has inspired several theoretical and practical traffic flow studies, the captured traffic states are limited to congested conditions, the time-space scope is also limited, and trajectory reconstruction methods used required many manual operations to gain high-fidelity data.

Recently, Unmanned Aerial Vehicle (UAV) based frameworks for high-resolution vehicle data collection have also been proposed (Barmpounakis and Geroliminis, 2020; Khan et al., 2017; Li et al., 2019) due to its comprehensive spatial coverage, flexibility, high resolution, and cost-effectiveness. (Khan et al., 2017), developed an automated five-staged methodological framework was developed for extracting multivehicle trajectories from UAV-acquired traffic footages. These steps include preprocessing of UAV-acquired videos, stabilization to minimize UAV instability, calibration to assign real-world distances and coordinates to the image coordinates, vehicle detection and tracking using background subtraction detection algorithm and the Lucas-Kanade optical flow tracking algorithm respectively, and lastly trajectory management. Although this study demonstrated reasonably good performance, the vehicle detection and tracking algorithm used was not robust enough, therefore, false detections and the effect of partial occlusion make the extracted trajectory noisy.

A study conducted by (Barmpounakis and Geroliminis, 2020)resulted in the first high-resolution traffic data acquired from drones. The study deployed a swarm of 10 UAVs hovering over the Athens Central Business District were used to record traffic streams in a congested 1.3km$^2$ area with more than 100km – lanes of the road network. About half a million trajectories of vehicles have been made available to the research community for studying traffic congestion. In another related study, (Krajewski et al., 2018) developed the HighD dataset obtained using UAV hovering over German highways to automatically extract 45,000km of naturalistic driving behavior from 16.5h of recorded video. U-Net - a common neural network architecture - was used for vehicle detection of drone-captured 4K (4096x2160) resolution videos. Vehicle trajectories were extracted using spatio-temporal neighbor clustering techniques to track and assign unique identifiers to individual vehicles. Additional trajectory postprocessing was done to retrieve smooth positions, speeds, accelerations. (Li et al., 2019) also used a similar pipeline but added compensated for motion between UAV images and registered images to enable pixel-actual distance mapping for vehicle speed estimation. Despite these benefits, drone-based aerial data collection is constrained by flight time due to limited battery capacity. Therefore, UAVs will not be practical for continuous high-resolution traffic data collection.

Inspired by the NGSIM study, the current paper presents a three-stage framework that can automatically extract high-resolution traffic data from surveillance cameras. The study integrates advanced object detection and tracking with a 2D homography technique; transforming distorted camera perspectives into bird-eye views to allow for accurate, continuous, high resolution traffic data. Surveillance cameras are used because they already exist and have wide deployment across most highways in the US. Additionally, since these cameras are fixed, it enables continuous data collection and monitoring along specific road corridors. A deep learning-based, state-of-the-art object recognition system was trained using a large database of labelled data that includes six vehicle categories and captures different lighting and weather conditions. To improve vehicle recognition accuracy and transferability across multiple cameras, novel data augmentation was introduced during training and inferencing. A combination of feature-based, spatio-temporal trackers are developed to construct unique trajectories for each unique vehicle as they travers the traffic scene. Individual trajectories are fed into an automatic camera calibration system which uses a two-layer matrix system to transform image coordinate scales into actual real-world coordinate scales. Transformed trajectories are finally used to lane-by-lane, class-by-class counts, speeds, gaps, and acceleration parameters automatically.

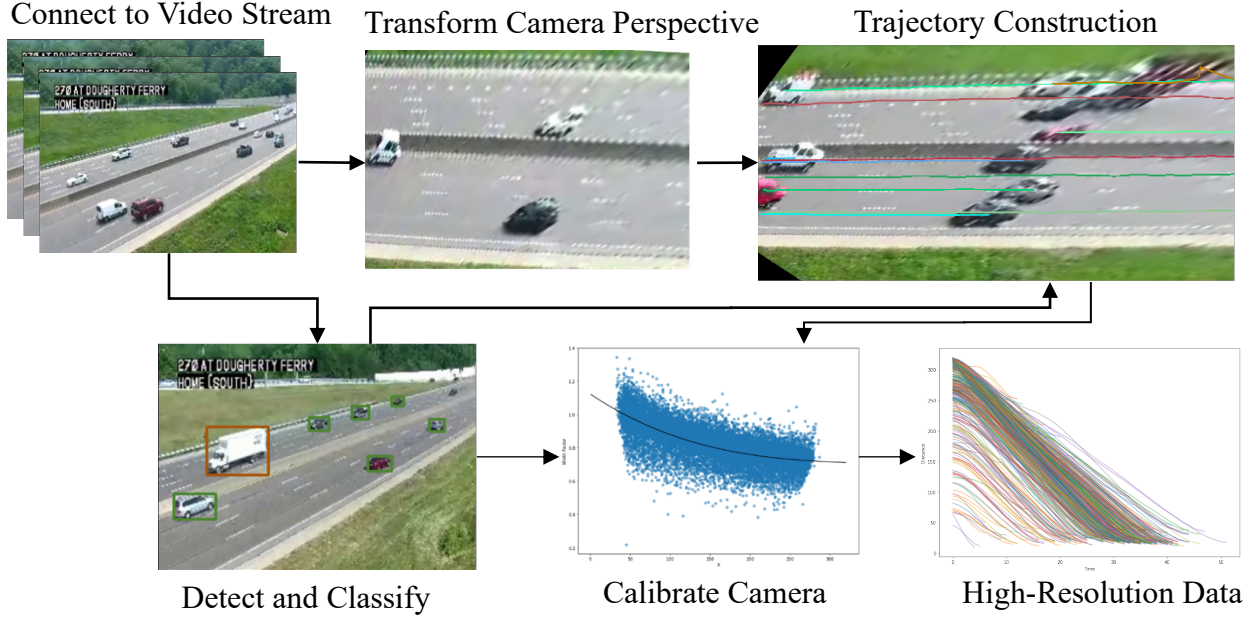

**Figure 1 Three-stage framework**

The current study improves the state-of-the-art with three key contributions:

1) Implements a 2D homography technique for perspective transformation of traffic scenes to a bird-eye view which reduces the effect of partial occlusion and improves the accuracy of speed and acceleration data collection.
2) CCTV cameras are automatically calibrated to convert pixel distances to real-world distances.
3) Used the concept of spatial-temporal neighbors to improve the tracking results and thereby address the challenge posed by misassignment of vehicle trajectories.

The outline of this paper is organized as follows. In Section 2, we discuss the study route, the surveillance cameras and details on the video streams collected. The methodology used to develop the vision system is discussed in Section 3. A detailed explanation of the three-stage framework is provided in the section. Section 4 summarizes the results of the proposed framework. Comparisons with ground truth data obtained from probed vehicles are used to validate the accuracy of the framework. Concluding remarks and future work are included in section 4.

## METHODOLOGY

This paper implements a three-stage framework to extract high-resolution traffic flow parameters from infrastructure-mounted surveillance cameras. The first stage is the functional stage which involves training and deploying a state-of-the-art deep learning model for object recognition. Next, we leverage computer vision techniques such as 2D homography to project the original perspective of CCTV cameras and vehicle detections into a birds-eye-view. This reduces the effect of camera perspective distortion and partial occlusions on traffic flow variables estimated. Then, object tracking algorithms are developed to associate each vehicle with a unique identifier, enabling us to construct trajectories for each vehicle. The final stage involves camera auto-calibration and the extraction of high-resolution traffic data from vehicle trajectories.

**Study Area and Video Data Collection**

The primary source of data used in the current study were traffic surveillance videos and images – for model training and probe data for evaluation of the accuracy of the vision system developed. The traffic images and videos were obtained from multiple surveillance cameras installed on I-270 in the St Louis district of Missouri, with a resolution of $320\times 240$ and a frame rate of 15 FPS. To train and generate robust models, datasets pertaining to different weather conditions were collected. Recording videos at the start of every hour for five minutes enables us to generate a diverse dataset. The probe data is obtained from HERE technologies. It provides high resolution aggregate speeds of vehicles on road segments. This data is used as the ground truth to evaluate the accuracy of speeds estimated from the vision-based system. **Table 1** shows an example of the input dataset.

**TABLE 1. Examples of Input Data**

| Video | FPS | Total Frames | Resolution | Source | Class Included |
|-------|-----|--------------|------------|--------|----------------|
| I270_Doughert | 15 | 4444 | $320 \times 240$ | MoDOT | 6 |
| I270 Clayton | 15 | 4185 | $320 \times 240$ | MoDOT | 7 |
| I270 at 364 | 15 | 4320 | $320 \times 240$ | MoDOT | 5 |
| I270 Gravois | 15 | 4153 | $320 \times 240$ | MoDOT | 7 |
| I270 Manchester | 15 | 4176 | $352 \times 240$ | MoDOT | 7 |
| I270 McKelvey | 30 | 8799 | $800 \times 450$ | MoDOT | 5 |

**Functional Stage**

*Object Recognition*

Yolov5, a state-of-the-art object detection model was used to detect and classify vehicles. The architecture of Yolov5 is shown in **Figure 2**. There are several reasons for selecting it for object detection. The Yolov5 architecture consists of three main blocks: a darknet backbone, neck and head. By incorporating a cross-stage partial network (CSPNet) (Wang et al., 2020) into the Darknet backbone, it is able to solve the problem of repeated gradient information and thereby decreasing the number of parameters of the model which improves the speed and accuracy of the model. Also, the Yolov5 applied path aggregation network (PANet) (Wang et al., 2019) as its neck to boost information flow by improving the propagation of low-level features. Additionally, adaptive feature pooling, which links feature grid and all feature levels, is used to make useful information in each feature level propagate directly to the following subnetwork. Lastly, the prediction vectors at the head of the Yolov5 architecture generate feature maps at 3 different sizes: 18 by 18, 36 by 36, and 2 by 72. As a result, it achieves multiscale prediction enabling it to detect small, medium, and large objects.

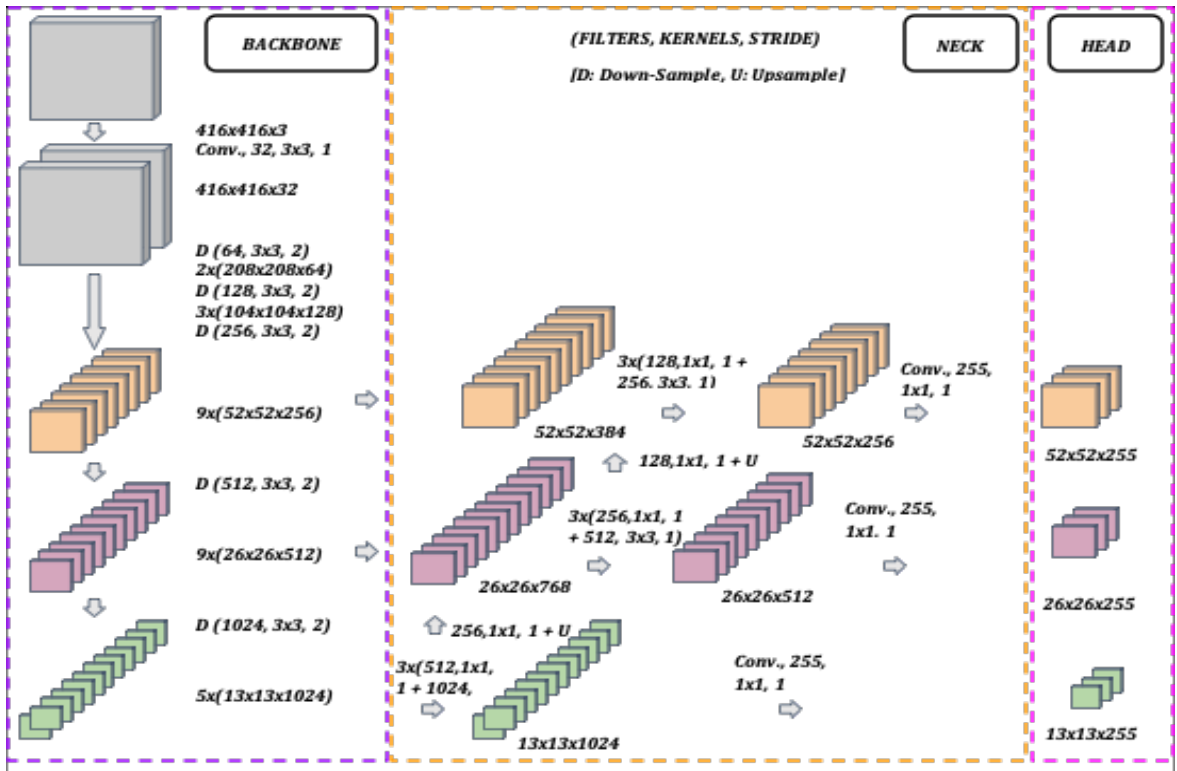

**Figure 2 Yolov5 structure**

Model Training: The object detection model was trained on a large database of annotations. About 80k images were manually annotated for different classes of vehicles including motorcycles, buses, single unit trucks, single trailer, and multi-trailer trucks. The model was trained on an NVIDIA GTX 2080Ti GPU. Details of our model training strategy and hyperparameter settings are shown in **Table 2**.

**TABLE 2 Hyperparameter Settings for Model Training**

| Model | Training Data | Test Data | Optimizer | LR | Batch | Epoch |
|-------|---------------|-----------|-----------|-----|-------|-------|
| Yolov5 | 81239 | 22,000 | SGD (Goyal et al., 2017)(Goyal et al., 2017)(21)(Goyal et al., 2017)(21) | $1 \times 10^{-3}$, $1 \times 10^{-2}$ | 16 | 400 |

To adapt Yolov5 to the current task, two main modifications were made to the default implementation: transfer learning and data augmentation.

Transfer Learning: This is implemented to enable the model to learn quickly while using less resources for training. Transfer learning could reduce the accuracy of the final model; however, it helps to avoid overfitting and improves the generalization of the trained model. In the current study, we froze the first 9 layers of the backbone by setting their gradients to zero prior to training. The remaining weights in subsequent layers were then used to compute the loss and updated by an optimizer.

Data Augmentation: Data augmentation during model training is crucial to avoid overfitting – a situation in which the network memorizes the training data features instead of learning them: this results in the model's accuracy on the unseen data (testing data) being drastically lower than training data (Zhang et al., 2021). The following data transformers were used: *Mosaic* - allows models to learn how to detect objects at smaller scales by combining four training images into one in certain ratios. This augmentation technique also significantly reduces the need for a large mini-batch size. *Scaling* - zooms in/out of the original image. *Flipping* - Flips the images horizontally and or vertically. Bounding Box Safe Crop - this method crop the image in a way that the bounding boxes will not remove from the original image. Additionally, test-time augmentation (TTA) was applied during inferencing. TTA involves using models to make predictions on multiple augmented copies of each image in the test set, and then returning an ensemble of the output predictions. TTA gives the model the best opportunity to correctly classify test images.

The F-1 scores reported in **Table 3** evaluate the performance of the vehicle recognition model developed. In general, motorcycles were the hardest to recognize. This is likely due to the mounting height and low resolution of surveillance cameras. Data augmentation during training did not yield any significant increase in F-1 scores. This might be due to the fact that the model has enough training data with minimal gaps to learn from. In fact, in some cases training data augmentation yielded a slightly low F1 score. Test time augmentation, however, significantly improved the performance of the model significantly. A visual evaluation of detection results under different scenarios is shown in **Figure 3**.

**TABLE 3 F-1 Scores for Different Object Classes**

| Vehicle Class (percentage of training data) | F 1 scores | | |
|---|---|---|---|
| | Without augmentation | Training data augmentation | Test-time augmentation |
| Motorcycle (3%) | 0.643 | 0.732 | 0.765 |
| Car (40%) | 0.698 | 0.721 | 0.823 |
| Bus (12%) | 0.832 | 0.821 | 0.867 |
| Single unit (19%) | 0.675 | 0.712 | 0.786 |
| Trailer (21%) | 0.932 | 0.914 | 0.954 |
| Multi-trailer (5%) | 0.876 | 0.901 | 0.634 |

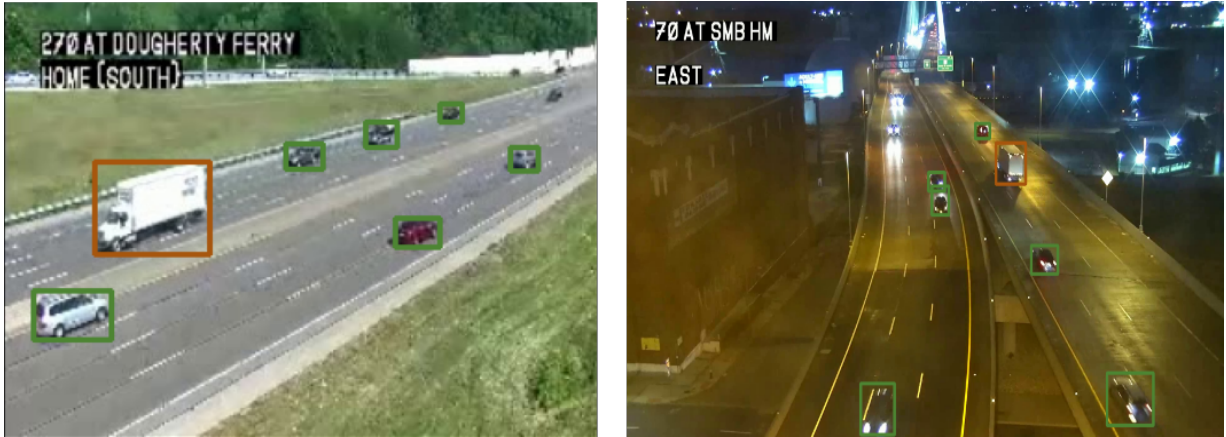

**Figure 3. Visual detection results in different scenarios**

**Object Tracking**

Object tracking associates each detected vehicle with a unique identifier enabling us to construct trajectories that can be used to generate high-resolution traffic data such as counts, speed, and acceleration. A key challenge to most object tracking algorithms that exist is vehicle occlusion. The current paper aims to overcome the impact of partial occlusion on tracking results by first detecting and automatically extracting the region of interest (ROI) from the original perspective of the video scene, inspired by the concept of two-point linear perspective. Subsequently, to mitigate the effects of partial occlusion, the paper corrects the original perspective view by transforming the ROI into a bird's-eye view using a perspective transformation matrix, which is also called Homography. This reduces the overlap and consequent misassignment between detected boxes.

*Region of Interest (ROI) Selection*

Drawing inspiration from the concept of the two-point perspective, consider a scenario depicted in **Figure 4(a)** where the viewer observes an object from a corner, similar to the view from the surveillance camera. In this situation, the parallel boundaries of the highway in the real world can be represented as two straight lines that converge towards a vanishing point within the camera frame. By introducing an additional vanishing point on the same horizon line, one can determine the corresponding points on both sides of the road boundaries.

Furthermore, by employing this process and utilizing the heuristic method, the second vanishing point is determined as the point located at a distance of twice the image width from the first vanishing point. In **Figure 4(b)**, the region of interest (ROI) of the road can be determined as shown in the blue box and subsequently utilized to be transformed into a bird-eye view.

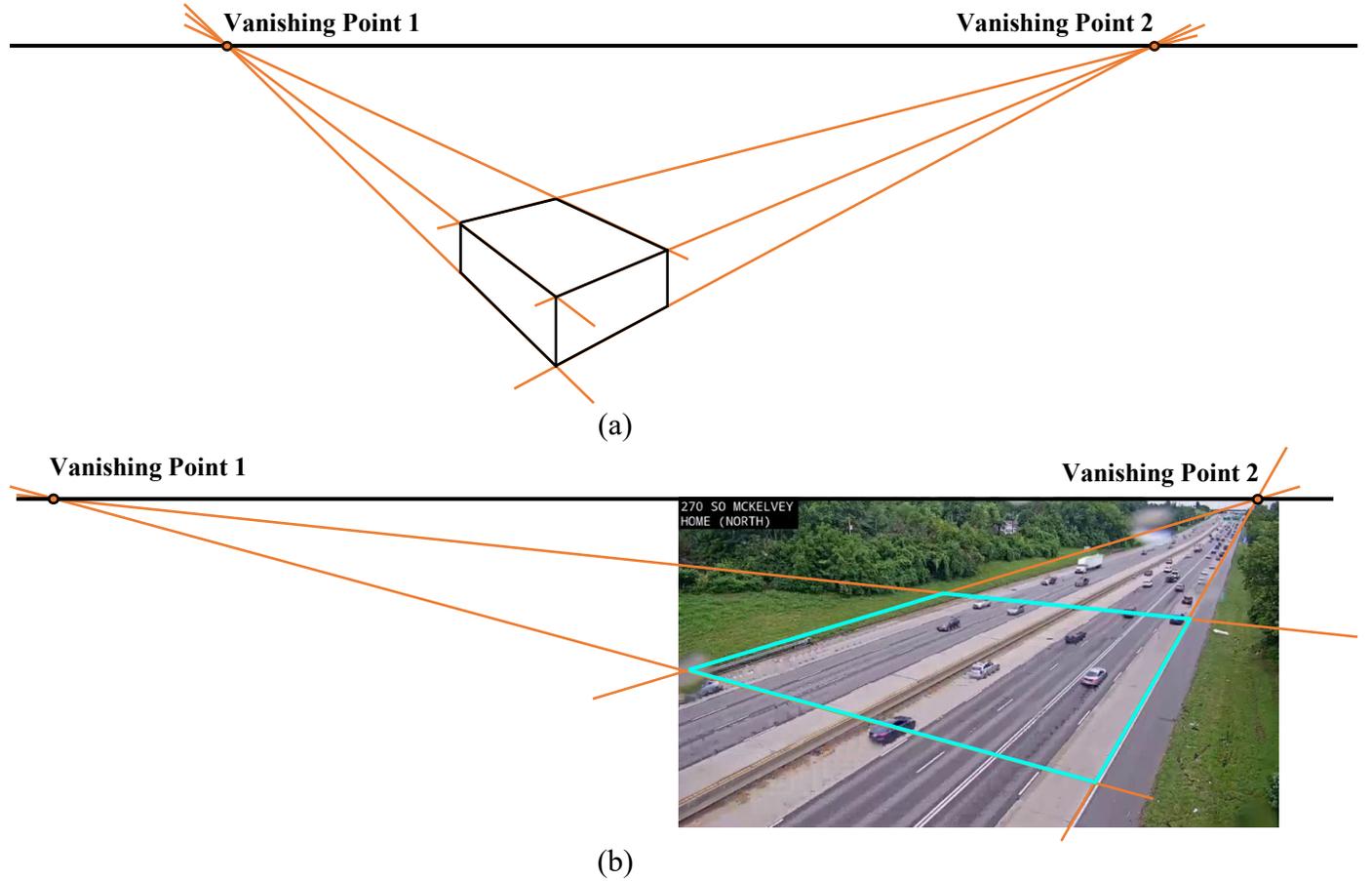

(a)

(b)

**Figure 4 Two-Point Linear Perspective Concept and Research of Interest (ROI) Selection**

*Homography Estimation*

The transformation matrix that maps the original scene to a top-down view is estimated via 2D homography. A system of homogenous coordinates is introduced so that all points can be handled correspondingly in projective geometry to convert points between 2D to 3D planes. Assume that a point in space is on a planar surface with known space and image plane coordinates. The homogenous coordinates of the corresponding 2D point $P = (X, Y)$ can be set as: $\overline{P} = \left(\overline{X}, \overline{Y}, \overline{Z}\right)^T$. As the third coordinate $Z \neq 0$, the point is equivalent to $P = \left(\frac{\overline{X}}{\overline{Z}}, \frac{\overline{Y}}{\overline{Z}}, 1\right)$.

Operating on these homogenous coordinates, any perspective transformation has the form as follows:

$$x_d = \overline{H}x_s \qquad (1)$$

Where $\overline{H}$ is the $3 \times 3$ perspective matrix or a reversible homogenous linear transformation on the projective plane. Thus, the **Equation 2** can also be written as:

$$\begin{bmatrix} x_d \\ y_d \\ 1 \end{bmatrix} \equiv \begin{bmatrix} \tilde{x}_d \\ \tilde{y}_d \\ \tilde{z}_d \end{bmatrix} = \begin{bmatrix} h_{11} & h_{12} & h_{13} \\ h_{21} & h_{22} & h_{23} \\ h_{31} & h_{32} & h_{33} \end{bmatrix} \begin{bmatrix} x_s \\ y_s \\ 1 \end{bmatrix} \qquad (2)$$

With 9 unknown parameters in the Homography, the degree of freedom of the matrix $\overline{H}$ is 8. This means at least four matched pairs of points, with at least three of them being non-collinear, is needed to compute the Homography matrix.

Applying direct linear transformation to the equations above, any given matched pair of points can be designated as:

$$x_d = \frac{h_{11}x_s + h_{12}y_s + h_{13}}{h_{31}x_s + h_{32}y_s + h_{33}}$$

$$(3)$$

$$y_d = \frac{h_{21}x_s + h_{22}y_s + h_{23}}{h_{31}x_s + h_{32}y_s + h_{33}}$$

$$(4)$$

**Equation 5** and **Equation 6** can also be written as follows for any given pair of points:

$$x_d(h_{31}x_s + h_{32}y_s + h_{33}) = h_{11}x_s + h_{12}y_s + h_{13}$$

$$(5)$$

$$x_d(h_{31}x_s + h_{32}y_s + h_{33}) = h_{11}x_s + h_{12}y_s + h_{13}$$

$$(6)$$

With all pairs of points, we have:

$$\begin{bmatrix} x_s^{(1)} & y_s^{(1)} & 1 & 0 & 0 & 0 & -x_d^{(1)}x_s^{(1)} & -x_d^{(1)}y_s^{(1)} & -x_d^{(1)} \\ 0 & 0 & 0 & x_s^{(1)} & y_s^{(1)} & 1 & -y_d^{(1)}x_s^{(1)} & -y_d^{(1)}y_s^{(1)} & -y_d^{(1)} \\ & & & & \vdots & & & & \\ x_s^{(n)} & y_s^{(n)} & 1 & 0 & 0 & 0 & -x_d^{(n)}x_s^{(n)} & -x_d^{(n)}y_s^{(n)} & -x_d^{(n)} \\ 0 & 0 & 0 & x_s^{(n)} & y_s^{(n)} & 1 & -y_d^{(n)}x_s^{(n)} & -y_d^{(n)}y_s^{(n)} & -y_d^{(n)} \end{bmatrix} \begin{bmatrix} h_{11} \\ h_{12} \\ h_{13} \\ h_{21} \\ h_{22} \\ h_{23} \\ h_{31} \\ h_{32} \\ h_{33} \end{bmatrix} = \begin{bmatrix} 0 \\ 0 \\ \vdots \\ 0 \\ 0 \end{bmatrix} \qquad (7)$$

**Equation 7** can be written in another form of $Ah = 0$. By computing the singular value decomposition (SVD) of the matrix $A$, $UDV^T = A$, the Homograph is obtained from the smallest singular value of vector V.

Since at least four matched pairs of points are needed to compute the Homography, four points that compose the ROI in the original perspective view, and the corresponding four points in the top-down view are selected to compute the unique Homography of each scenario. Consider **Figure 4** (a) for example, the four points- A, B, C, and D- selected in the original perspective view are along with the road, which is assumed to be straight. Thus, the straight-line AB and CD, which are parallel with the road are acceptable to use. Knowing the corresponding 4 points $A \to A'$, $B \to B'$, $C \to C'$ and $D \to D'$, the estimated transformation matrix will be used to warp the detected center coordinates of each object from object recognition models in the original perspective view to the top-down view shown in the second row of **Figure 4**. It is important to notice from the figure the significant decrease overlap between vehicles in the projected birds-eye view. This increases the accuracy of object tracking algorithms as the effect of occlusion is reduced. Additionally, in the projected view, lane widths do not vary with distance substantially as compared to the original view. This is crucial for speed and acceleration estimation.

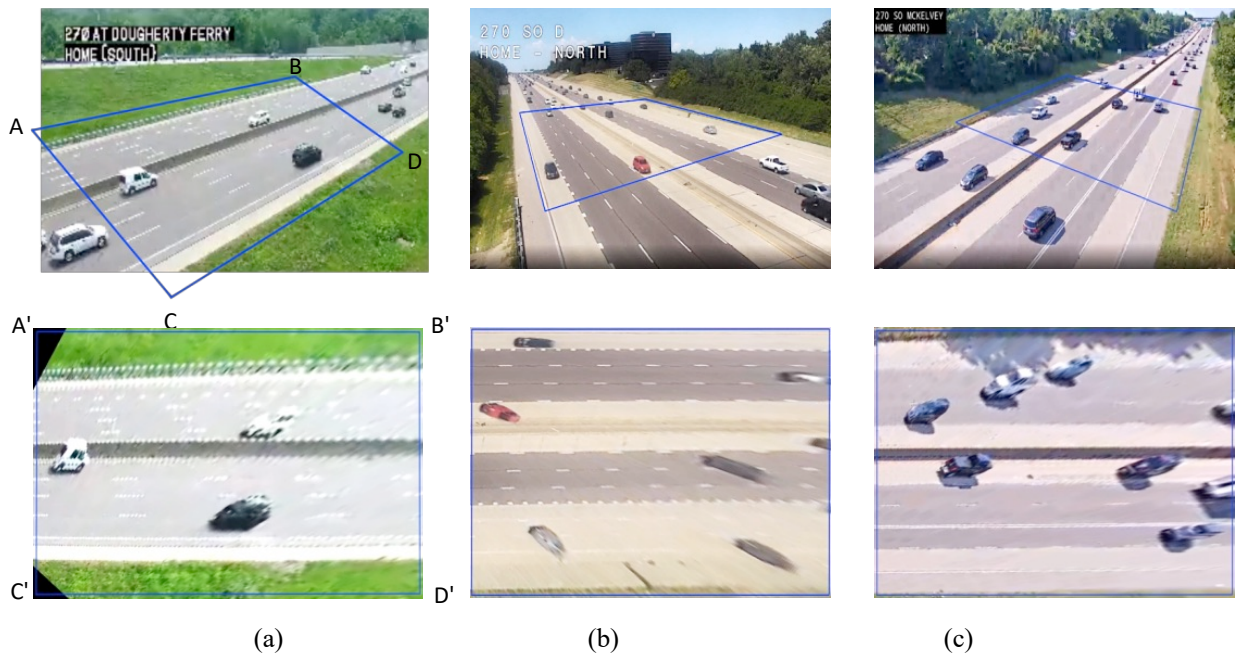

(a)                           (b)                           (c)

**Figure 4 Centroid of vehicles within selected area in original view and top-down view**

*Vehicle Trajectory Construction*

Vehicle tracking is carried out in the projected top-down view by using two tracking techniques- Motpy (Bewley et al., 2016), and a simple multi-object tracker, and BYTEtrack (Zhang et al., 2022), a prevalent multi-object tracking method. Motpy employs motion prediction, along with a Kalman Filter, feature similarity matching, and an IOU based cost metric, to associate the detected objects and assign them to appropriate tracks. The algorithm in **Table 4** presents the key steps of Motpy used to track objects in the current study. BYTEtrack is an effective and straightforward technique for object tracking that proficiently handles occlusion and enhances tracking performance. Rather than discarding bounding boxes with low confidence scores, BYTEtrack preserves all bounding boxes and leverages the overlap between the detection box and the track. This enables the extraction of occluded objects from the low-scoring boxes using Kalman Filter and an IOU-based cost metric, thereby maintaining the continuity of the track. Additionally, BYTEtrack exclusively relies on detected bounding boxes, allowing for both speed and effectiveness in its operation. The core tracking procedure of BYTEtrack, known as BYTE (Zhang et al., 2022), is presented in **Table 5.**

In addition, both tracking techniques are equipped with new bounding boxes that are estimated using the average size of objects observed in the top-down view for each class. By incorporating these new bounding boxes, both Motpy and BYTEtrack efficiently operate on the transformed bird-eye view, ensuring effective tracking performance.

**TABLE 4 Algorithm of Motpy Tracking**

| **Algorithm** Motpy Tracking | |
|---|---|
| 1: **function** TRACKER($\boldsymbol{detections}, \boldsymbol{\sigma_l}, \boldsymbol{\sigma_h}, \boldsymbol{\sigma_{iou}}, \boldsymbol{min_{tsize}}$) | |
| 2:     let $\boldsymbol{\sigma_l}$ ← low detection threshold | |
| 3:     let $\boldsymbol{\sigma_h}$ ← high detection threshold | |
| 4:     let $\boldsymbol{\sigma_{iou}}$ ← IOU threshold | |
| 5:     let $\boldsymbol{min_{tsize}}$ ← minimum track size in frames | |
| 6:     let $\boldsymbol{T_a}$ ← [] | active tracks |
| 7:     let $\boldsymbol{T_f}$ ← [] | finished tracks |
| 8:     **for** $\boldsymbol{frame}, \boldsymbol{dets}$ in $\boldsymbol{detections}$ **do** | |
| 9:       $\boldsymbol{dets}$ ← filter for $\boldsymbol{dets}$ with $\boldsymbol{scores} \geq \boldsymbol{\sigma_l}$ | |
| 10:      let $\boldsymbol{T_u}$ ← [] | updated tracks |
| 11:      **for** $\boldsymbol{t_i}$ in $\boldsymbol{T_a}$ **do** | |
| 12:         **if** $\boldsymbol{not\ empty}(\boldsymbol{dets})$ **then** | |
| 13:           $\boldsymbol{b_{iou}}, \boldsymbol{b_{cost}}$ ← cost_matrix_iou_feature($\boldsymbol{t_i}, \boldsymbol{dets}$, | |
| 14:                 **feature_similaritu_fn**($\boldsymbol{t_i}, \boldsymbol{dets}$), feature_similaritu_($\boldsymbol{t_i}, \boldsymbol{dets}$)) | |
| 15:           **if** $\boldsymbol{b_{iou}}[[r,c]\ \boldsymbol{for}\ [r,c]\ \boldsymbol{in}\ \boldsymbol{b_{cost}}] \geq \boldsymbol{\sigma_{iou}}$ **then** | |
| 16:             append_new_detection($\boldsymbol{t_i}, [\boldsymbol{r,c}]$) | |
| 17:             set_max_score($\boldsymbol{t_i}$, box_score([$\boldsymbol{r,c}$])) | |
| 18:             set_class($\boldsymbol{t_i}$, box_class([$\boldsymbol{r,c}$])) | |
| 19:             $\boldsymbol{T_u}$ ← append($\boldsymbol{T_u}, \boldsymbol{t_i}$) | |
| 20:             remove($\boldsymbol{dets}, [\boldsymbol{r,c}]$) | remove box from $\boldsymbol{dets}$ |
| 21:         **if** $\boldsymbol{empty}(\boldsymbol{T_u})$ or $\boldsymbol{t_i}$ is_not $\boldsymbol{last}(\boldsymbol{T_u})$ **then** | |
| 22:           **if** get_max_score($\boldsymbol{t_i}$) $\geq \boldsymbol{\sigma_h}$ or size($\boldsymbol{t_i}$) $\geq \boldsymbol{min_{tsize}}$ **then** | |

23:           $T_f \leftarrow \text{append}(T_f, t_i)$
24:      $T_f \leftarrow$ new tracks from $\boldsymbol{dets}$
25:      $T_a \leftarrow T_u + T_n$
25:   return $T_f$

**TABLE 5 Algorithm of BYTE**

| Algorithm Pseudo-code of BYTE |
|---|

**Input:** A video sequence V; object detector Det; detection score threshold $\boldsymbol{\tau}$
**Output:** Tracks $\boldsymbol{T}$ of the video

1:    Initialization: $T \leftarrow \emptyset$
2:    **for** $\boldsymbol{frame\ f_k\ in}$ **V do**
         / * Figure 2 (a) */
         / * predict detection boxes & scores */
3:       $\boldsymbol{D_k} \leftarrow \text{Det}(\boldsymbol{f_k})$
4:       $D_{high} \leftarrow \emptyset$
5:       $\boldsymbol{D_{low}} \leftarrow \emptyset$
6:       **for** $\boldsymbol{d\ in\ D_k}$ **do**
7:          **if** $\boldsymbol{d.score} > \boldsymbol{\tau}$ **then**
8:            $\boldsymbol{D_{high}} \leftarrow \boldsymbol{D_{high}} \cup \{\boldsymbol{d}\}$
9:          **end**
10:        **else**
11:           $\boldsymbol{D_{low}} \leftarrow \boldsymbol{D_{low}} \cup \{\boldsymbol{d}\}$
12:        **end**
13:      **end**

         / * predict new locations of tracks */
14:       **for** $t\ in\ T$ **do**
15:         $\boldsymbol{t} \leftarrow$ **KalmanFilter**$(\boldsymbol{t})$
16:      **end**
         / * Figure 2 (b) */
         / * first association */
17:      Associate $\boldsymbol{T}$ and $\boldsymbol{D_{high}}$ using Similarity #1
18:      $D_{remain} \leftarrow$ remaining object boxes from $\boldsymbol{D_{high}}$
19:      $T_{remain} \leftarrow$ remaining tracks from $\boldsymbol{T}$

         / * Figure 2 (c) */
         / * second association */
20:      Associate $T_{remain}$ and $\boldsymbol{D_{low}}$ using Similarity #2
21:      $T_{re-remain} \leftarrow$ remaining tracks from $T_{remain}$

         / * delete unmatched tracks */
22:      $T \leftarrow T \setminus T_{re-remain}$

         / * initialize new tracks */
23:      **for** $\boldsymbol{d\ in}\ D_{remain}$ **do**

24:  | | $T \leftarrow T \cup \{d\}$
25:  | **end**
26:  **end**
27:  Return: $\boldsymbol{T}$

*Note.* Reprinted from (Zhang et al., 2022)

**Figure 5** shows tracking results across different cameras.

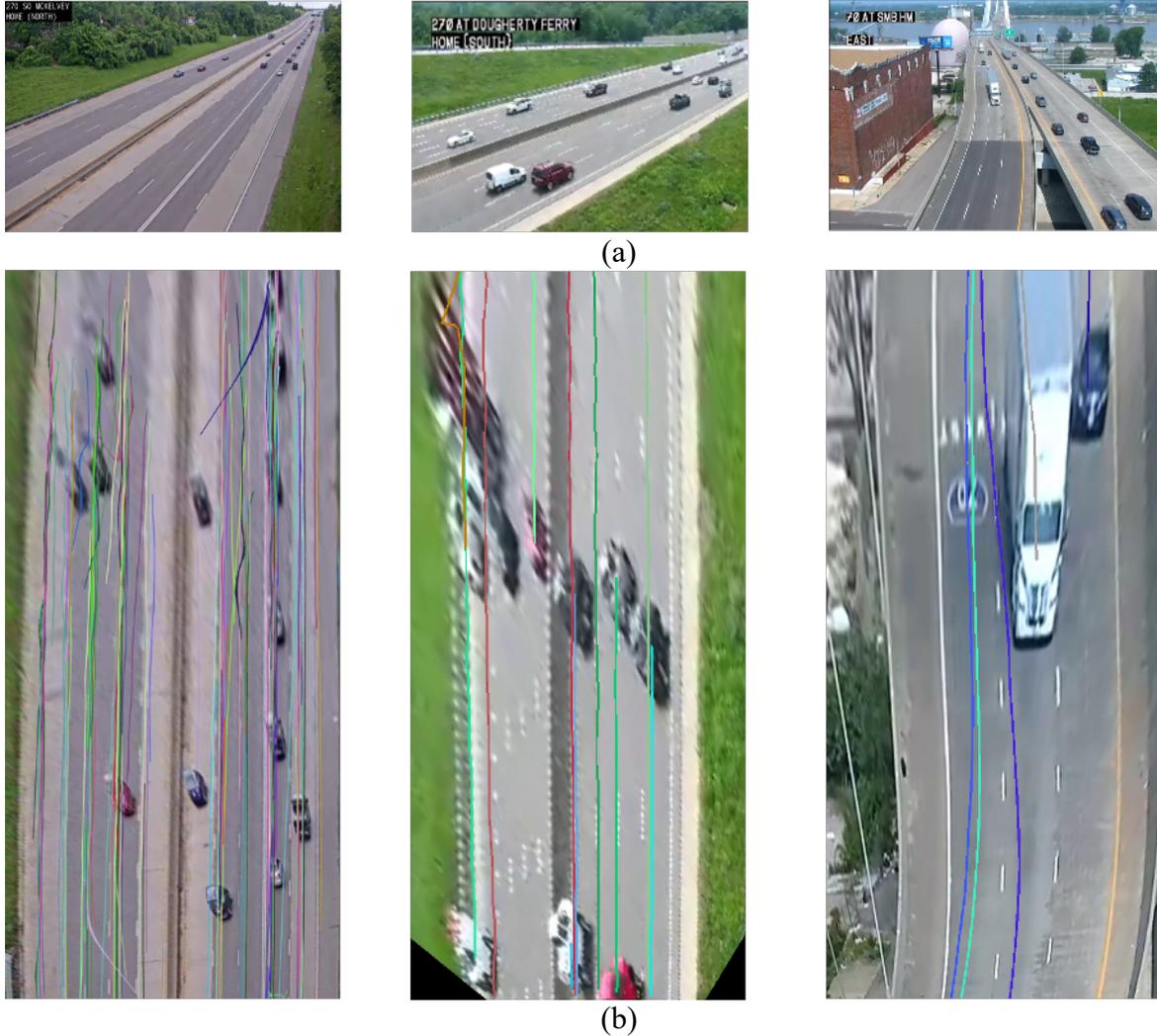

(a)

(b)

**Figure 5 Vehicles in original view with detection result (b) Vehicles in projected top-down view with tracking result**

*Post Processing of Fragmented Tracks*

Factors such as occlusion and missed object detections can lead to fragmented trajectories. To obtain accurate high-resolution data, the fragmented trajectories (smoothed) are post-processed to re-connect them by means of spatio-temporal neighbors clustering. We first build a *begin* and *end* structure for all trajectories. Let's call these structures fragments. A fragment can have a kind: *b* for *begin* and *e* for *end* . It has a reference to the detection at that location, its track id, frame, and a simplification (piecewise linear segments) of the track. Looping through each frame of video, at frame $N$ , track ($T_i$) is about to exit (reached its *end* fragment at frame $N$ ). We try to find a join-able track for $T_i$. using the following algorithm:

1. Temporal neighbors: Find the $begin$ of fragments that are ahead of $T_i$. by $\frac{fps}{1sec}$. These are joinable temporal neighbors.
2. Find spatial neighbors among the temporal neighbors. A spatial neighbor is when the $begin$ fragments starting coordinate is in the bounding box of $end$ fragment
    a. The spatial neighbor should also have the same direction as $T_i$
3. Compute direction deflection metrics between $end$ fragments and their spatial neighbors, and filter neighbors based on directional heuristics – keep neighbors with deflection $\leq 30^o$.
4. Perform candidate selection using filtered results from step 3
    a. For each $T_i$ feature points of the vehicle on frame $N$ are extracted using the Harris Corners and tracked by Optical Flow on the specific frame from $N + 1$ to $N + fps$ .
    b. The fragment is considered as a joinable trajectory if the predicted position is close enough to the beginning of the fragment, $T_i$ . The threshold of the distance between the prediction position of $T_i$ and the beginning of the fragment is based on the image resolution.

Results of applying the trajectory reconstruction algorithm are shown in **Figure 6**. As observed, the majority of fragmented trajectories have been joined together resulting in much longer trajectories which will improve the accuracy of counts, speed and acceleration information that will be extracted.

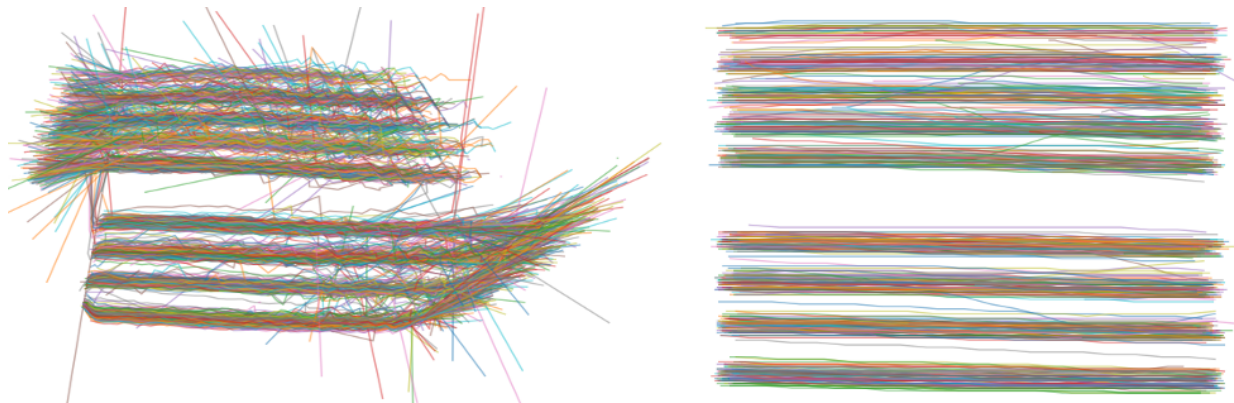

**Figure 6 Trajectories colored by unique identification before and after post-processing**
*Direction*
Since the study was carried out on State highways, there were usually two directions of traffic which were assigned by computing bearings from the start and end coordinates of each trajectory. Individual lanes of travel were identified by projecting the trajectories along the horizontal and vertical axis of the projected top-down view image. The results in **Figure 7** (a) and (b) show how all lanes in both directions in the projected top-down view image shown were detected with this approach.

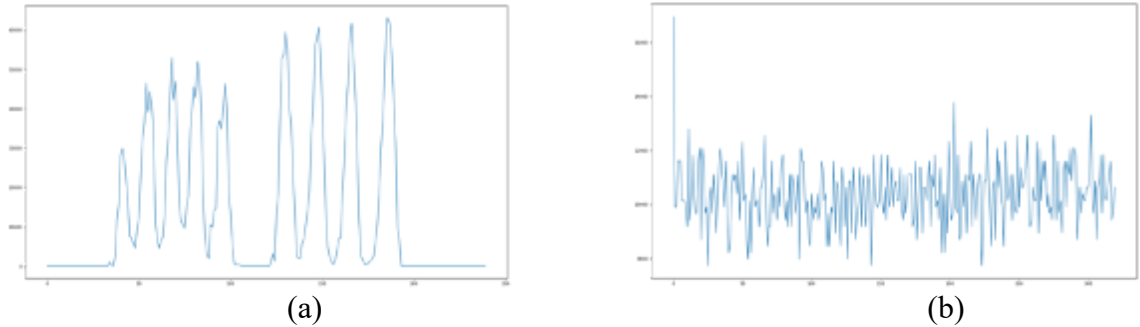

(a)                                                           (b)

**Figure 7 (a) Histogram plot of raw data in vertical form. (b) in horizontal form**

**High-Resolution Parameter Estimation**

To extract high-resolution data, each camera needs to be calibrated so that pixel distances can be mapped directly to actual real-world distances. In this study, the calibration was automatically performed based on a 2-layer matrix approach described as follows:

*Camera Auto-Calibration*

For each scene, we track the height and width of bounding boxes as a vehicle with respect to their actual dimensions as it traverses the scene. In general, the widths and heights of the bounding boxes increase in length (in the projected view) as the vehicle moves away from the camera. To correct for this discrepancy, we developed a 2-layer matrix (one for width and the other for height) that uses a linear regression model to learn the relationship between the bounding box width-height and the image resolution along the x and y dimensions respectively. The outcome of this model is used as a correction factor that is applied at the pixel level to convert image distances to real-world values. **Figure 8** below shows the correction factor that was learned to correct pixel distances measured from the camera site shown. The heatmap shows the magnitude of correction applied per location of vehicle. It is important to note that although the trend of these correction factors is generally similar for different cameras, their magnitudes may change, hence there is a need to re-calibrate for each camera.

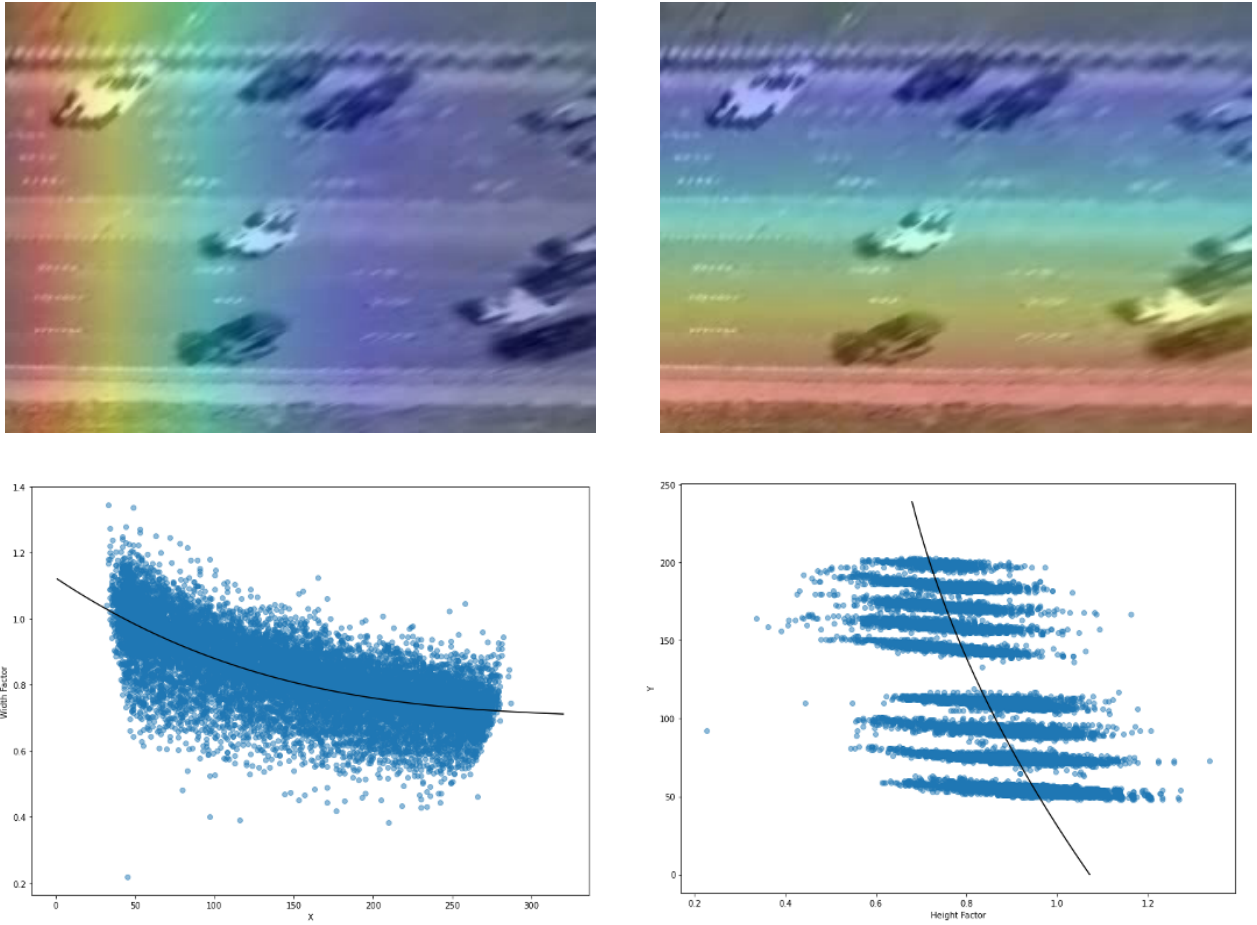

**Figure 8.** Top Row: Heatmap of 2-layer matrix used for calibrating measured pixel distances. Bottom Row: Learned correction factor used to calibrate a camera

*Speed and Acceleration Estimation*

The speed of each vehicle at a specific time $t$ is calculated by determining the calibrated distance traveled as shown in **Equation 8**.

$$V = \frac{D}{T}$$

$$\tag{8}$$

$$D = \sqrt{(dx)^2 + (dy)^2}$$

$$\tag{9}$$

$$dx = 14.7 \cdot k_w \cdot dx_p \ and \ dy = 6 \cdot k_h \cdot dy_p$$

$$\tag{10}$$

In **Equation 10**, 14.7 ft and 6 ft are the average length and height dimensions of a sedan. $k_w$ and $k_h$ are calibration constants that are used to correct pixel movements in the image width ($dx_p$) and height ($dy_p$) dimensions. The derivative of vehicle speeds estimated over 5 second period is used to calculate acceleration. In addition to individual vehicle speed estimation, the average speed over space was also computed to enable comparison with ground truth probe vehicle speed data. The space mean speed was estimated using **Equation 11**:

$$v_t = \frac{N}{\sum_{n=1}^{N} \frac{1}{v_n}}$$

(11)

**Figure 9** below shows lane by lane speed estimated over a period of 5 minutes. As expected, the inner lanes have higher speeds compared to the outer lanes for both directions. There is also a speed gradient from one lane to the other which is consistent with ground truth observations and traffic flow theory. Histogram plots in **Figure 9** show the distribution of speeds and accelerations over the period of analysis for the camera location shown.

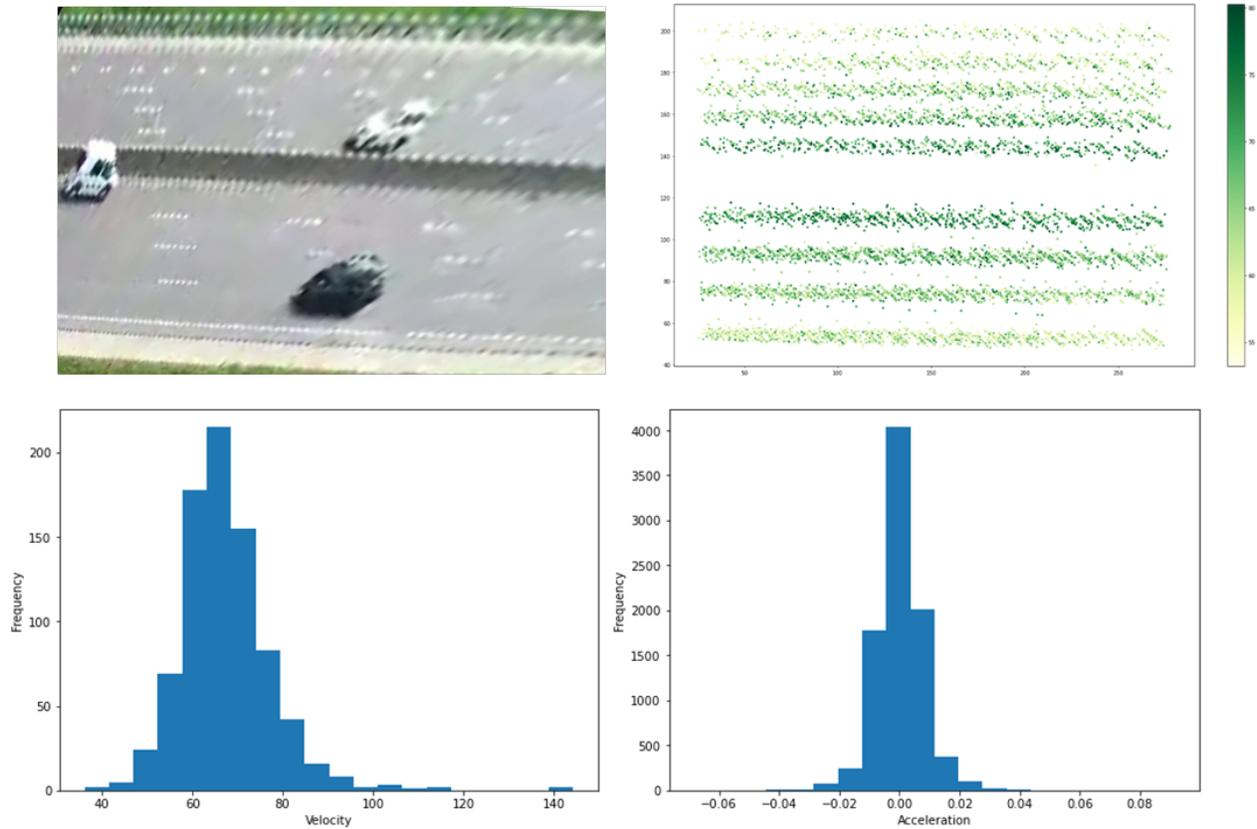

**Figure 9 First row: lane by lane heatmap for individual vehicle trajectories. Second row: distribution of speeds and acceleration**

**RESULTS AND DISCUSSIONS**

       This section evaluates the performance of the vision system across 4 different cameras with varying levels of congestion at different times of the day. **Table 6** and **Table 7** show the results of the high-resolution data from different videos. From these two tables, by given a specific calibration factor, the proposed three-stage framework can obtain the high-resolution data from videos. **Table 6** summaries the counting results for each camera location based on direction and vehicle classification in both perspective view and top-down view. Also shown in **Table 6** is the comparison results of counting in original perspective view and top-down view using Motpy and BYTEtrack tracking techniques respectively. On the other hand, **Table 7** presents the results of speed and acceleration. There was no benchmark data to compare vehicle accelerations, so only results from the vision system were reported. Also, the probe data used as the speed benchmark does not report speed by vehicle classification, as observed speeds are not compared across vehicle class.

       *Directional Counts*: Counts from the vision system were compared against manual counts over the same duration. There is a significant improvement in the accuracy of counting results achieved by the BYTEtrack tracking technique compared to the Motpy tracking technique. In general, during free flow conditions, traffic counting error rates were between 1% and 5%. The average accuracy of the vision system drops around 93.6% for segment I-270/Dougherty and I-270/364, and 96% - 98% for segment I-270/Clayton and I-270/Gravious in the top-down view when using BYTEtrack. It is obvious that the counting results obtained from the top-down view are more precise than those from the perspective view, indicating the effectiveness of perspective transformation techniques in mitigating the issue of partial occlusion. With the exception of I-270/Gravois, the other three locations exhibit lower accuracy for trucks and semi-trailers, when compared to cars. This suggests that gathering more samples of trucks and semi-trailers would be beneficial in improving the framework's performance. Additionally, because of processing counts in a birds-eye view, there were no significant differences in accuracy of counts by direction. Directions closer and those farther from the camera had similar accuracies.

       *Speed Estimation*: Speed estimates from the vision system were compared to vehicle probe data speeds for the same period of time. The table indicates < 10% absolute error between the speeds estimated by the camera and the that from probe. Specifically, the error for cars is 5.02% and for trucks, it is 15.32%. The largest differences were observed between the minimum and maximum speeds. The length of road segment over which probes speeds are calculated is much longer than the field of view of the camera. This difference could be the reason for the differences in min-max speeds.

**TABLE 6 Result of the High-Resolution Traffic Data – Counts**

| Segment | Class | D* | Real Counts | Top- down View Fine-tuning YOLOv5 | | | | Perspective View Fine-tuning YOLOv5 | | | |
|---|---|---|---|---|---|---|---|---|---|---|---|
| | | | | Motpy | | BYTE truck | | Motpy | | BYTE truck | |
| | | | | Counts | ER% | Counts | ER% | Counts | ER% | Counts | ER% |
| I270 Dougherty | Car | 1 | 548 | 525 | 4.20 | 542 | 1.09 | 611 | 11.50 | 654 | 19.34 |
| | | 2 | 440 | 416 | 5.45 | 442 | 0.45 | 562 | 27.73 | 631 | 43.41 |
| | Truck | 1 | 32 | 21 | 34.38 | 29 | 9.38 | 67 | 109.38 | 58 | 81.25 |
| | | 2 | 7 | 5 | 28.57 | 8 | 14.29 | 50 | 614.29 | 18 | 157.14 |
| I270 Clayton | Car | 1 | 332 | 284 | 14.46 | 335 | 0.90 | 547 | 64.76 | 414 | 24.70 |
| | | 2 | 472 | 452 | 4.24 | 462 | 2.12 | 604 | 27.97 | 580 | 22.88 |
| | Truck | 1 | 18 | 14 | 22.22 | 19 | 5.56 | 40 | 122.22 | 35 | 94.44 |
| | | 2 | 29 | 27 | 6.90 | 31 | 6.90 | 54 | 86.21 | 47 | 62.07 |
| I270 364 | Car | 1 | 303 | 290 | 4.29 | 302 | 0.33 | 372 | 22.77 | 384 | 26.73 |
| | | 2 | 320 | 303 | 5.31 | 312 | 2.50 | 418 | 30.63 | 403 | 25.94 |
| | Truck | 1 | 10 | 11 | 10.00 | 11 | 10.00 | 37 | 270.00 | 25 | 150.00 |
| | | 2 | 16 | 11 | 31.25 | 18 | 12.50 | 42 | 162.50 | 34 | 112.50 |
| I270 Gravois | Car | 1 | 439 | 444 | 1.14 | 450 | 2.51 | 535 | 21.87 | 614 | 39.86 |
| | | 2 | 330 | 311 | 5.76 | 345 | 4.55 | 410 | 24.24 | 415 | 25.76 |
| | Truck | 1 | 26 | 17 | 34.62 | 27 | 3.85 | 34 | 30.77 | 30 | 15.38 |
| | | 2 | 23 | 13 | 43.48 | 23 | 0.00 | 21 | 8.70 | 23 | 0.00 |

* D indicates the direction

**TABLE 7 Result of the High-Resolution Traffic Data – Speed and Acceleration**

| Segment | Class | D* | Speed from Probe (MPH) | | | | | Estimated Speed (MPH) | | | | | Estimated Acceleration(m/s²) | |
|---|---|---|---|---|---|---|---|---|---|---|---|---|---|---|
| | | | Min | 25% | 50% | 75% | Max | Min | 25% | 50% | 75% | Max | Mean | 50% |
| I270 Dougherty | Car | 1 | 63 | 64 | 65 | 65 | 66 | 53 | 62 | 66 | 70 | 79 | 0.0023 | 0 |
| | | 2 | 56 | 60 | 64 | 64 | 66 | 53 | 60 | 65 | 69 | 79 | -0.0002 | 0 |
| | Truck | 1 | 63 | 64 | 65 | 65 | 66 | 45 | 45 | 45 | 51 | 56 | 0.0089 | 0.009 |
| | | 2 | 56 | 60 | 64 | 64 | 66 | 66 | 66 | 66 | 75 | 75 | -0.0076 | -0.007 |
| I270 Clayton | Car | 1 | 14 | 15 | 15 | 16 | 17 | 9 | 13 | 15 | 20 | 29 | 0 | 0 |
| | | 2 | 56 | 58 | 60 | 61 | 63 | 42 | 50 | 54 | 57 | 62 | -0.0009 | 0 |
| | Truck | 1 | 14 | 15 | 15 | 16 | 17 | 9 | 9 | 10 | 15 | 22 | 0 | 0 |
| | | 2 | 56 | 58 | 60 | 61 | 63 | 69 | 69 | 69 | 69 | 69 | -0.0001 | 0 |
| I270 364 | Car | 1 | 49 | 51 | 53 | 56 | 58 | 35 | 40 | 45 | 51 | 69 | 0.0003 | 0 |
| | | 2 | 62 | 63 | 63 | 63 | 65 | 46 | 59 | 63 | 67 | 72 | -0.0004 | 0 |
| | Truck | 1 | 49 | 51 | 53 | 56 | 58 | 44 | 44 | 48 | 48 | 48 | -0.0001 | 0 |
| | | 2 | 62 | 63 | 63 | 63 | 65 | 53 | 53 | 55 | 55 | 55 | -0.0005 | 0 |
| I270 Gravois | Car | 1 | 31 | 32 | 34 | 37 | 42 | 22 | 33 | 37 | 43 | 51 | 0.0004 | 0.001 |
| | | 2 | 56 | 60 | 64 | 65 | 66 | 50 | 56 | 62 | 68 | 76 | -0.0013 | -0.002 |
| | Truck | 1 | 31 | 32 | 34 | 37 | 42 | 20 | 30 | 35 | 40 | 44 | 0.0009 | 0 |
| | | 2 | 56 | 60 | 64 | 65 | 66 | - | - | - | - | - | - | - |

## CONCLUDING REMARKS

This paper implements a vision system for extracting high resolution traffic data from infrastructure-mounted surveillance cameras. Using the architecture of a state-of-the-art deep learning model, Yolov5, we trained a model to detect and classify 6 types of vehicles. Transfer learning and data augmentation techniques were used to reduce overfitting and increase the generalization of the developed model. The principles of 2D homography are used to correct camera perspective distortions by transforming the original camera view to a birds-eye view. Vehicle trajectories are subsequently constructed in the birds-eye view using two prevalent tracking techniques, that combines with IOU cost functions and spatio-temporal neighborhood clustering methods. Finally, a 2-layer matrix approach is used to automatically calibrate each camera enabling us to transform pixel distances to real-world distances.

The framework developed was used to extract traffic counts, speed, and acceleration data at 4 locations with varying levels of congestion. The results showed about +/- 5.4% average error rate for directional traffic counts during free flow, and +/- 3.2% average error rate during congestion. Speed bias between camera estimates and probe data estimates remained below 10% MSE although there were significant differences between minimum and maximum speeds. The vision system was also able to accurately capture lane specific data; able to distinguish between high speed and low speed lanes. The accuracy of the results shown can be attributed to high object detection rates, reduced levels of occlusion due to tracking in the projected bird-eye view and automatic camera calibration.

Although the calibration approach used in the current framework was robust, there were discrepancies when the field of view of the camera was significantly reduced. This affected speed and acceleration estimates from the models developed. Future directions of this study will look into monocular depth estimation techniques to derive vehicle speeds. This will eliminate the need for calibration and hopefully improve the models' generalization.

## AUTHOR CONTRIBUTIONS

The authors confirm contribution to the paper as follows: study conception and design: Adu-Gyamfi; data collection: Zhang; model training: Adu-Gyamfi; data analysis and interpretation of result: Zhang, Mussah, Yu, Adu-Gyamfi; draft manuscript: Zhang, Mussah, Adu-Gyamfi. All Author reviewed the results and approved the final version of the manuscript.